\documentclass[runningheads]{llncs}

\usepackage{eccv}

\usepackage{eccvabbrv}

\usepackage{graphicx}
\usepackage{booktabs}

\usepackage[accsupp]{axessibility}  

\usepackage{hyperref}

\usepackage{orcidlink}

\usepackage{bbm}

\begin{document}

\title{OmniPoser: Flexible Human Motion Recovery in the Wild with Masked Flow Matching} 

\titlerunning{OmniPoser}

\author{Minghao Liu\inst{1,2}\orcidlink{0009-0007-5758-2717} \and
Tutian Tang\inst{1,2}\orcidlink{0000-0001-8254-2038}}

\authorrunning{M. Liu and T. Tang}

\institute{Shanghai Jiao Tong University, China \and
Shanghai Artificial Intelligence Laboratory, China\\
\email{\{lmh209,tttang\}@sjtu.edu.cn}}

\maketitle

\begin{abstract}
  Recovering 3D human motion from sparse wearable sensors is challenging
  due to the inherent noise, occlusion, and instability of real-world
  tracking signals.
    Existing approaches are restricted to fixed sensor configurations and,
  when built on diffusion models, are computationally prohibitive for
  real-time deployment.
    We propose OmniPoser, a universal framework that recovers full-body SMPL
  motion from a flexible combination of heterogeneous inputs, such as 3D keypoints,
  sparse IMU rotations, head-mounted 6-DoF poses, or any subset
  thereof, within a single model.
    OmniPoser introduces a cross-modal masking mechanism paired with a Masked
  Conditioning Encoder, enabling one architecture to incorporate diverse sensor
  configurations through a masked design.
    A decoupled dual-stream generative backbone generates complete motion via Conditional Flow Matching, requiring
  only a single ODE integration step at inference.
    Across three settings and seven benchmarks, including in-the-wild Nymeria
  and Ego-Exo4D dataset, OmniPoser performs on par with state-of-the-art methods
  while running at over 700\,FPS.
    The project page is available at
  \url{https://jamesbridgewater51.github.io/omniposer-eccv/}.
  
\end{abstract}

\begin{figure}
    \centering
    \includegraphics[width=1\linewidth]{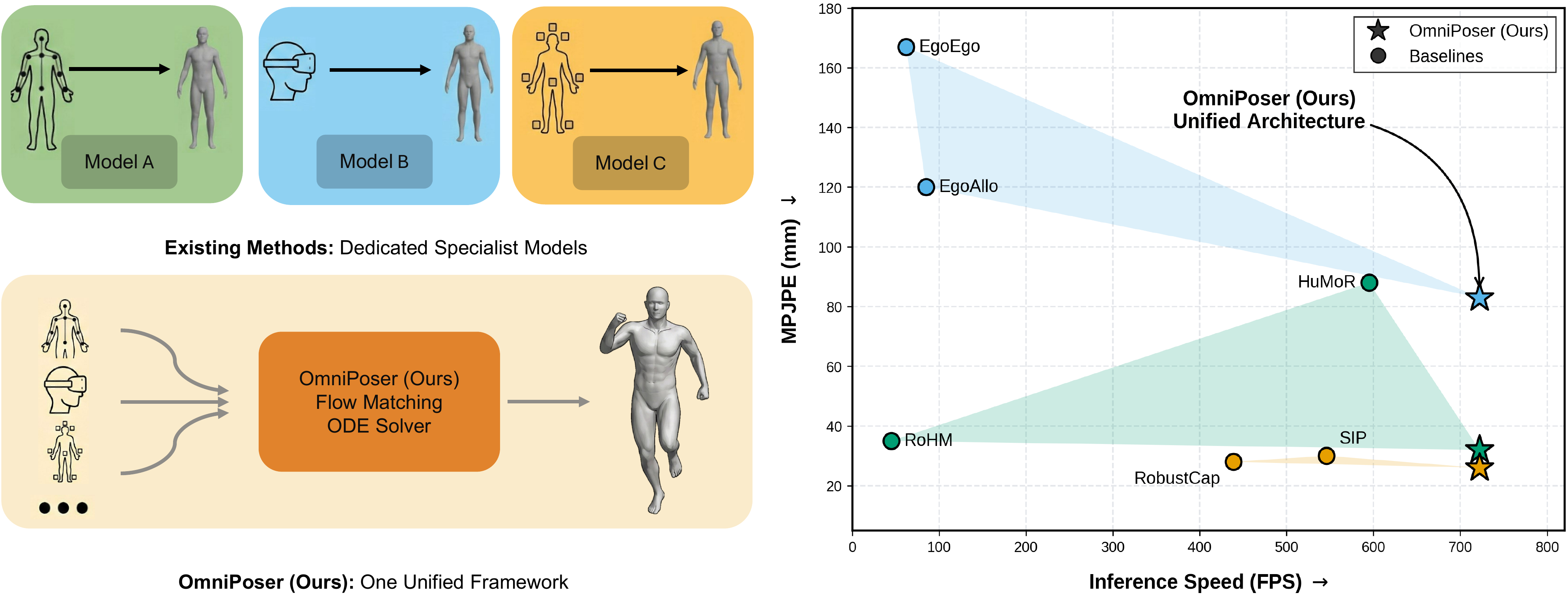}
\caption{\textbf{From dedicated specialist models to one unified framework.}
\emph{Left:} Existing approaches require a dedicated architecture for each sensor configuration (e.g. partial-body keypoints (RoHM), head-mounted 6-DoF (EgoAllo), sparse IMUs (SIP/RobustCap)), making every new modality a re-engineering effort.
OmniPoser incorporates all of them with one flow-matching backbone and training recipe, recovering full-body SMPL motion from \emph{different} sensor configurations via the msaked design.
\emph{Right:} Accuracy--FPS performance across three settings.
Each baseline (circle) is plotted; OmniPoser (star) matches or surpasses baseline methods at $>$700\,FPS with lower Mean Per Joint Position Error (MPJPE).
}

    \label{fig:teaser}
\end{figure}

\section{Introduction}
\label{sec:intro}

The rapid advancement of AR/VR, immersive gaming, and human-robot interaction has driven a high demand for accurate yet real-time human motion capture. While traditional motion recovery is well-solved in heavily constrained studio environments using dense optical markers, consumer applications require motion capture to operate seamlessly in the wild, relying on limited and imperfect input signals.
These intermediate signals can take various forms depending on the hardware ecosystem: 3D keypoints from monocular vision~\cite{Zhang2024RoHMRH,Rempe2021HuMoR3H}, rotations from wearable IMUs~\cite{pan2023fusing,tang2026stereoinertialposermetricaccurateshapeaware} or smartwatches~\cite{lee2024mocap}, or 6D poses from Head-Mounted Displays (HMDs) and hand-held joysticks~\cite{winkler2022questsim,jiang2022avatarposer}. Crucially, in real-world scenarios, all these signals are inevitably noisy, fragmented, and unstable. Vision-based 3D keypoints are often corrupted by occlusion and lighting conditions, while rotation readings from sparse sensors may drift by nature or even disconnect due to wireless interference.

To recover human motion in the wild with such inputs,
previous works have transitioned from deterministic inverse kinematics regressors~\cite{li2021hybrik,jiang2024manikin} to data-driven generative models. Notably, diffusion models~\cite{Wouwe2023DiffusionPoserRH,Zhang2024RoHMRH} have been employed to iteratively denoise and infer plausible poses under severe occlusions or sparse node conditions.

However, these methods expose two limitations.
First, they are typically designed for a predefined configuration of inputs—such as a fixed combination of keypoints or rotations from a specific set of sensors.
This inflexibility requires the networks to be entirely re-trained for different sensor configurations, and also makes them difficult to handle the unstable nature of real-world signals.
Second, they are usually too computationally expensive and thus unsuitable for real-time applications on consumer-grade mobile devices.

In this work, we propose OmniPoser, a universal framework that recovers human motion from a flexible combination of noisy and unstable signals.
OmniPoser employs a Conditional Flow Matching backbone to enable highly efficient motion generation. In order to support flexible inputs, we introduce a cross-modal masking mechanism that unifies heterogeneous sensor configurations within one model. Zeroing out missing entries, however, destroys the identity of unobserved joints and leads to poor recovery (see ablation in \cref{subsec:ablation}). We therefore design a Masked Conditioning Encoder (MCE) that replaces each missing joint with a learnable per-joint embedding, allowing the network to reason about \emph{which} joints are absent and recover them accordingly. At inference, a single Euler integration step maps Gaussian noise onto the motion manifold conditioned on the masked observations, enabling real-time motion recovery.

We evaluate OmniPoser across three different settings, each targeting a very distinct sensor configuration, and compare against state-of-the-art methods. Altogether we evaluate our method quantitatively on five benchmarks~\cite{Mahmood2019AMASSAO,ma2024nymeria,li2021aistplusplus,huang2022rich,trumble2017totalcapture}.
Results show our method performs on par with state-of-the-art methods, while maintaining a processing speed over 700 FPS, as illustrated in \cref{fig:teaser}.

The main contributions of this paper can be summarized as:

(1) We propose OmniPoser, a Conditional Flow Matching-based framework for human motion recovery that accepts heterogeneous sparse observations, such as 3D keypoint positions, joint rotations, and their combinations, as input.

(2) We introduce a cross-modal masking mechanism paired with a Masked Conditioning Encoder (MCE) that replaces unobserved joints with learnable per-joint embeddings, enabling the network to distinguish which joints are missing rather than treating all of them identically.

(3) We demonstrate competitive performance compared to state-of-the-art methods and real-time inference capabilities across diverse configurations and highly challenging in-the-wild datasets.

\section{Related Work}
Our method is closely related to human motion capture with sparse sensors, general human motion recovery, and Flow Matching models for human kinematics.

\textbf{Human Motion Capture with Sparse Sensors}.
Wearable human motion capture systems offer a portable alternative to constrained optical systems in studios.
Existing approaches generally fall into three hardware paradigms: IMU-based systems, head-mounted cameras, and VR devices with sparse sensors.
(1) \textbf{IMU-based Systems:} While commercial solutions require dense sensor arrays, recent research successfully reconstructs motion from sparse setups, such as 6 IMUs (e.g., TransPose~\cite{Yi2021TransPose}, PIP~\cite{Yi2022PhysicalIP} and PNP~\cite{Yi2022PhysicalIP}) or even as few as 4 IMUs (e.g., LIP~\cite{ren2023lidar}). 
Under fixed sensor placements, SAIP~\cite{Yin2025SAIP} incorporates shape-aware motion tracking tailored to real body shape, and GlobalPose~\cite{Yi2025GlobalPose} utilizes physics priors to refine global tracking. For a more flexible paradigm, generative frameworks like DiffusionPoser~\cite{Wouwe2023DiffusionPoserRH} achieve similar body motion recovery while supporting a flexible combinations of IMU signals. 
  Conversely, to fundamentally counteract inherent integration drift via explicit geometric guidance, inertial-visual fusion~\cite{pan2023fusing} remains essential.
 Most recently, Stereo-Inertial Poser~\cite{tang2026stereoinertialposermetricaccurateshapeaware} has been proposed to utilize a single stereo camera and six IMUs for metric-accurate, shape-aware motion capture. 
(2) \textbf{Head-Mounted Cameras (HMC):} Reconstructing full-body poses from egocentric vision suffers from severe lower-body self-occlusion. Systems like EgoEgo~\cite{Li2022EgoBodyPE} and EgoExo~\cite{li2021egoexo} utilize SLAM and hand poses to condition full-body generation. More recently, EgoAllo~\cite{Yi2024EstimatingBA} utilized allocentric grounding and conditional diffusion to eliminate floating artifacts, while HMD2~\cite{guzov2025hmd} further refined egocentric motion estimation.
(3) \textbf{VR Devices and Sparse Sensors:} Inside-out tracking in mixed reality typically uses an HMD and two controllers. Approaches like QuestSim~\cite{winkler2022questsim} use reinforcement learning to simulate physics-based full-body avatars from these inputs. AvatarPoser~\cite{jiang2022avatarposer}, EgoPoser~\cite{jiang2024egoposer}, and MANIKIN~\cite{jiang2024manikin} utilize deep kinematics and optimization for highly accurate tracking, occasionally fusing supplementary IMUs~\cite{dai2024hmdposer} or smartwatches~\cite{lee2024mocap}. Despite these advancements, the majority of these tracking methods remain structurally rigid and tailored to a single, fixed sensor configuration.
Despite these advancements, the majority of these methods remain structurally rigid and tailored to a single, fixed sensor configuration.

\textbf{Human Motion Recovery}.
Human motion recovery from monocular videos typically follows a two-stage pipeline. First, off-the-shelf pose detectors~\cite{cao2017realtime,wang2020deep,xu2022vitpose} extract spatial keypoints. Second, a sequence modeling network processes these spatiotemporal coordinates to compute 3D motion over time. Early generative approaches like HuMoR~\cite{Rempe2021HuMoR3H} utilized conditional variational autoencoders to optimize motion priors. Recently, Denoising Diffusion Probabilistic Models~\cite{ho2020denoising} have dominated this field. RoHM~\cite{Zhang2024RoHMRH} reconstructs globally coherent motions by decomposing the task into global and local diffusion models. It can solves different patterns of keypoints combinations (e.g., upper-body v.s. lower-body), so it serves as our primary baseline.
Multi-hypothesis frameworks like CHAMP~\cite{zhang2024champ} and Platypose~\cite{pierzchlewicz2024platypose} robustly handle uncertainty but they also lack topological flexibility.
To address in-the-wild complexities, HumanMM~\cite{zhang_humanmm_2025} uses multi-shot alignment to ensure global motion continuity across dynamic camera transitions. A recent concurrent work, MoRo~\cite{moro}, explores masked modeling to achieve occlusion-robust recovery from monocular inputs.
Overall, existing frameworks are insufficiently universal. With focus on monocular images, they heavily rely on the spatial keypoints produced by pose detectors in the first stage. In contrast, our proposed method natively infers human motion from an arbitrary combination of 3D keypoints and sparse IMU rotations.

\textbf{Masked Motion Modeling and Decomposition}.
Techniques like masked modeling and local-global motion decomposition have been iteratively refined into mature paradigms within the HMR community. Specifically, PromptHMR~\cite{Wang2025PromptHMR} leverages spatial and semantic prompts to robustly guide human shape recovery. MEGA~\cite{Fiche2025MEGA} addresses visual ambiguity by formulating mesh recovery as a feature completion task via a masked generative autoencoder. GENMO~\cite{Li2025GENMO} unifies generation and estimation by treating variable conditioning signals as explicitly masked inputs alongside a joint local-global motion representation. EgoMDM~\cite{Shin2026EgoMDM} embeds local-to-global translation constraints into a diffusion framework to synthesize motion from sparse egocentric sensors. Additionally, PUMA~\cite{Kim2026PUMA} optimizes the underlying generative training dynamics through progressive unmasking schedules. We embrace these established designs in our work, and adapt them to heterogeneous sparse-observation recovery. We expand the mask from a dropout training schedule into a masked module that unifies different sensor modalities, dictating whether a joint is observed via a 3D position, a rotation, a 6-DoF signal.

\section{Method}
\label{sec:method}

Recovering 3D human motion from sparse, heterogeneous, and occluded observations remains fundamentally ill-posed. Different sensor modalities (e.g. sparse IMU readings, head-mounted egocentric cameras, monocular RGB detectors, optical trackers, or partial motion capture) provide only a narrow and noisy window into the full-body state.
Our key insight is that regardless of sensor modality, all observations can be cast as \emph{partial views} of the same canonical body representation, and the completion of unobserved degrees of freedom can be formulated as a unified generative conditioning problem.
In the following, we first define the unified motion representation (\cref{subsec:representation}).
We then introduce the cross-modal masking mechanism and the Masked Conditioning Encoder (MCE) that converts masked observations into a joint-aware conditioning signal (\cref{subsec:masking}).
Finally, we present the Conditional Flow Matching formulation, the decoupled generative architecture, and the training objectives (\cref{subsec:generation}).

\begin{figure}[!t]
    \centering
    \includegraphics[width=1\linewidth]{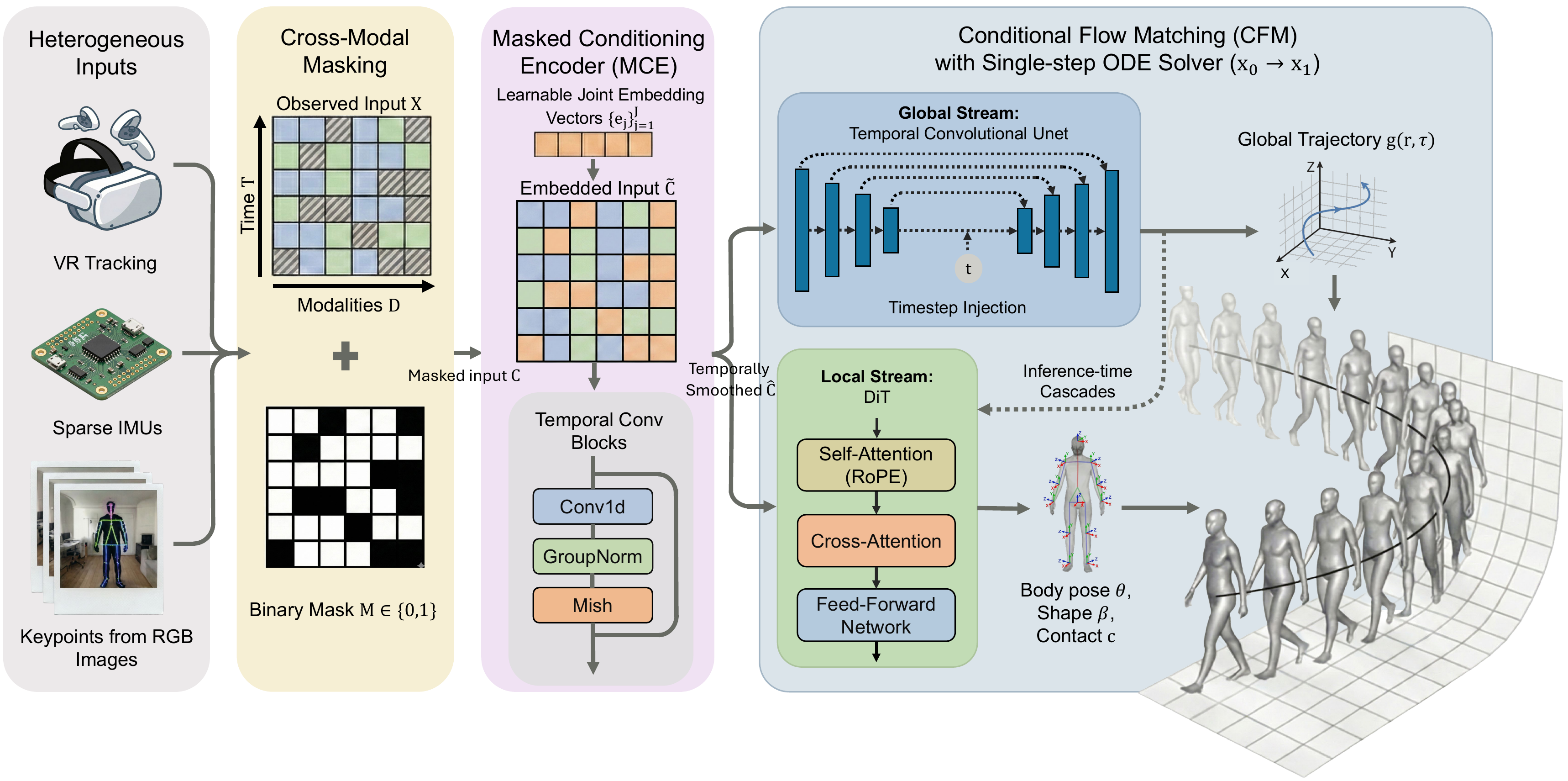}
    \vspace{-0.6cm}
    \caption{Overview of OmniPoser. Heterogeneous observations (multi-modal 3D keypoints and rotations) are unified through a cross-modal masking mechanism and processed by Masked Conditioning Encoder (MCE)~(\cref{subsec:masking}). A decoupled dual-stream architecture(\cref{subsec:architecture}) (a temporal convolutional UNet for global trajectory and a cross-attention Diffusion Transformer (DiT) for local body pose) generates SMPL parameters via Conditional Flow Matching(\cref{subsubsec:flow_matching}), achieving real-time inference in a single ODE step. Best viewed in color.}
    \label{fig:pipeline}
\end{figure}

\subsection{Motion Representation}
\label{subsec:representation}

%
We represent a motion clip of $T$ frames as a time series $\mathbf{X} \in \mathbb{R}^{T \times D}$, where each row $\mathbf{x}^{(t)} \in \mathbb{R}^{D}$ is a concatenation of semantically distinct \emph{feature groups}, describing a different aspect of the body state at time $t$. $D$ is the sum of each feature group's dimension.
In this work, we use feature groups based on the SMPL~\cite{loper2023smpl} model, including the global root trajectory (root translation $\boldsymbol{r}$ and root rotation $\boldsymbol{\tau}$), 3D joint positions $\mathbf{J} \in \mathbb{R}^{T \times J \times 3}$ and 6D rotations $\mathbf{R^w}$ or $\mathbf{R^r} \in \mathbb{R}^{T \times J \times 6}$ (in the world frame or in the root frame), body shape parameters $\boldsymbol{\beta} \in \mathbb{R}^{10}$, and binary foot contact labels $\mathbf{c}$. These features are concatenated to form the complete motion representation $\mathbf{x}^{(t)}$ at each frame, serving as both the partial input and the fully recovered output.
The full representation vector is thus $\mathbf{x}^{(t)} = [\boldsymbol{r}; \boldsymbol{\tau}; \mathrm{vec}(\mathbf{J}); \mathrm{vec}(\mathbf{R}^{\mathrm{w}}); \mathrm{vec}(\mathbf{R}^{\mathrm{r}}); \boldsymbol{\beta}; \mathbf{c}] \in \mathbb{R}^{D}$.
This representation serves as both input and output. On the input side, a subset of feature groups and joints are observed, depending on the sensor configuration. On the output side, the model recovers all feature groups for all joints, where joint rotations are expressed in the local (parent-relative) frame as the standard SMPL body pose.

\subsection{Cross-Modal Masking and Masked Conditioning Encoder}
\label{subsec:masking}

%
To unify heterogeneous sensor inputs within one model, we design a configurable \emph{mask scheme} that specifies, for a given sensor configuration, which feature groups and which joints are observed.
Concretely, a mask scheme defines:
(i)~a set of \emph{visible joint indices} $\mathcal{V} \subseteq \{1, \ldots, J\}$ controlling per-joint masking for \emph{joint-based} feature groups (e.g., local positions, per-joint rotations), and
(ii)~visibility flags for \emph{global} feature groups (e.g., whether root translation or orientation is observed).
Given a mask scheme, we construct a binary feature-level mask $\mathbf{M} \in \{0,1\}^{T \times D}$ that zeros out unobserved features in the conditioning tensor.
However, zero-padding removes the identity of missing joints; therefore, the network cannot distinguish \emph{which} joints are absent, leading to large reconstruction errors (\cref{subsec:ablation}). We address this with the Masked Conditioning Encoder described below.

For a feature group, the mask visibility may vary along the temporal axis, enabling the representation of intermittent sensor dropout, faithfully mimicking the signal interruptions common in real-world wearable or markerless capture systems.
This design enables one model to incorporate a wide range of input configurations, from optical motion capture ($|\mathcal{V}|{=}22$, rotations visible), to 5-point VR tracking ($\mathcal{V}{=}\{\text{head, hands, feet}\}$, positions only), to egocentric head-only estimation ($\mathcal{V}{=}\{\text{head}\}$) through our masked design.

During training, mask samples are drawn from  several configurations. This includes predefined sensor layouts such as VR/AR tracking, HMD with controllers, sparse keypoints from RGB images, and full IMU suits. It also includes data-driven visibility patterns: we estimate per-joint visibility probabilities from the large-scale, in-the-wild Ego-Exo4D dataset~\cite{Grauman2023EgoExo4DUS} using its COCO~\cite{coco} pose annotations predicted by MMPose~\cite{mmpose2020}. These real-world occlusion statistics mirror the inputs produced by upstream pose-estimation systems in practice. In \cref{subsec:qualitative}, we show that this training strategy enables OmniPoser to generalize to in-the-wild scenarios, including ego-centric activities from Ego-Exo4D that were unseen during training.

\subsubsection{Masked Conditioning Encoder}
\label{paragraph:MCE}
While the mask scheme defines \emph{what} is observed, the network must also know \emph{which specific joints} are missing in order to recover them. Zero-padding treats all absent joints identically, discarding their positional identity. To solve this, we introduce a \emph{Masked Conditioning Encoder} (MCE) that converts the raw masked conditioning signal $\mathbf{C} \in \mathbb{R}^{T \times D}$ into a joint-aware representation.
Inspired by Masked Autoencoders (MAE)~\cite{He2021MaskedAA}, the MCE maintains a bank of $J$ learnable \emph{joint embedding vectors} $\{\mathbf{e}_j\}_{j=1}^{J}$, each matching the per-joint feature dimension.
At each timestep $t$, unobserved entries are replaced by the corresponding learnable embedding:
\begin{equation}
    \tilde{\mathbf{c}}^{(t)}_j = 
    \begin{cases}
        \mathbf{c}^{(t)}_j & \text{if } j \in \mathcal{V}^{(t)}, \\
        \mathbf{e}_j      & \text{otherwise},
    \end{cases}
    \label{eq:mae_mask}
\end{equation}
where $\mathcal{V}^{(t)}$ denotes the set of visible joints at time $t$ (which may vary per timestep in probabilistic masking settings).
The masked-and-embedded tensor is then processed by a stack of $K=4$ residual temporal convolution blocks, each followed by Group Normalization and Mish activations, producing a temporally smoothed output $\hat{\mathbf{C}} \in \mathbb{R}^{T \times D}$ that serves as the conditioning signal for the generative backbone.
The learnable embeddings provide a fill-in signal richer than zero-padding. This design lets the model specialize during training to encode per-joint positional priors. Then temporal convolutions interpolate short-duration occlusion gaps before the conditioning enters the backbone.

\subsection{Generative Architecture and Training}
\label{subsec:generation}

\subsubsection{Conditional Flow Matching.}
\label{subsubsec:flow_matching}
The conditioned signal $\hat{\mathbf{C}}$ produced by the MCE is fed into a generative model based on Conditional Flow Matching (CFM)~\cite{Lipman2022FlowMF}.
Given a ground-truth sample $\mathbf{x}_1$ and Gaussian noise $\mathbf{x}_0 \sim \mathcal{N}(\mathbf{0}, \mathbf{I})$, a conditional optimal-transport (OT) affine path constructs intermediate states
\begin{equation}
    \mathbf{x}_t = \alpha_t \, \mathbf{x}_1 + \sigma_t \, \mathbf{x}_0,
    \label{eq:affine_path}
\end{equation}
where $(\alpha_t, \sigma_t)$ are the linear interpolation coefficients of the OT path.
A denoising network $v_\theta$ is trained to predict the clean target $\mathbf{x}_1$ from $\mathbf{x}_t$ and the conditioning $\hat{\mathbf{C}}$, supervised by the squared $\ell_2$ error $\| v_\theta(\mathbf{x}_t, t, \hat{\mathbf{C}}) - \mathbf{x}_1 \|_2^2$.
At inference, a single Euler integration step from $t{=}0$ to $t{=}1$ maps Gaussian noise onto the SMPL motion manifold.

\subsubsection{Decoupled Architecture}
\label{subsec:architecture}

A complete human motion can be split into global trajectory and local body pose, however, jointly predicting them from partially observed inputs is challenging because they live in different spaces: global trajectory is low-DoF (degrees of freedom) but must exhibit long-horizon coherence (drift-free), while local body pose is high-DoF but primarily needs to meet biomechanical constraints.
To address this, the denoising network $v_\theta$ are two streams that are trained independently yet share the same MCE conditioning interface: a \emph{global stream} that generates the root orientation $\boldsymbol{r}$ and translation $\boldsymbol{\tau}$, and a \emph{local stream} that generates body pose $\boldsymbol{\theta}$, shape $\boldsymbol{\beta}$, and foot contact $\mathbf{c}$.
Each stream's backbone is chosen to match its output characteristics: a convolutional UNet with multi-scale skip connections for the low-dimensional global trajectory, and a Transformer-based Diffusion Transformer (DiT) employing Rotary Position Embedding (RoPE)~\cite{su2024roformer} and cross-attention to $\hat{\mathbf{C}}$ for the high-dimensional local body pose.

During training, the local stream receives ground-truth trajectory as part of its conditioning. At inference, the two streams are composed: the global stream's predicted trajectory is fed into the local stream as conditioning, enabling cascaded generation where global trajectory informs local pose recovery.

\subsubsection{Training Objectives}
\label{subsubsec:losses}
\label{subsec:training}

Beyond the flow matching loss $\mathcal{L}_{\mathrm{FM}}$, we include auxiliary losses that operate on the decoded SMPL parameters.
The predicted motion representation $\hat{\mathbf{X}}$ is passed through SMPL forward kinematics to give reconstructed 3D joint positions $\hat{\mathbf{J}} \in \mathbb{R}^{T \times J \times 3}$.
We summarize the loss items as follows. For detailed definitions, please refer to the supplementary materials.

\paragraph{Local Stream Losses.}
The local stream training objective combines flow matching with representation, kinematic, and physics-grounded regularization:
\begin{equation}
\begin{split}
    \mathcal{L}_{\mathrm{pose}} = {} & \mathcal{L}_{\mathrm{FM}} + \lambda_{\mathrm{repr}} \mathcal{L}_{\mathrm{repr}} + \lambda_{\mathrm{jpos}} \mathcal{L}_{\mathrm{jpos}} \\
    & + \lambda_{\mathrm{vjpos}} \mathcal{L}_{\mathrm{vjpos}} + \lambda_{\mathrm{jvel}} \mathcal{L}_{\mathrm{jvel}} + \lambda_{\mathrm{smooth}} \mathcal{L}_{\mathrm{smooth}} \\
    & + \lambda_{\mathrm{fc}} \mathcal{L}_{\mathrm{fc}} + \lambda_{\mathrm{skate}} \mathcal{L}_{\mathrm{skate}}
\end{split}
\label{eq:pose_loss}
\end{equation}
Here $\mathcal{L}_{\mathrm{repr}}$ penalizes representation error; $\mathcal{L}_{\mathrm{jpos}} {=} \| \hat{\mathbf{J}} - \mathbf{J}^{\mathrm{GT}} \|_2^2$ and $\mathcal{L}_{\mathrm{jvel}} {=} \| \Delta_t \hat{\mathbf{J}} - \Delta_t \mathbf{J}^{\mathrm{GT}} \|_2^2$ supervise joint positions and velocities after SMPL forward kinematics; $\mathcal{L}_{\mathrm{smooth}} {=} \| \Delta_t^2 \hat{\mathbf{J}} \|_2^2$ penalizes acceleration for temporal smoothness; $\mathcal{L}_{\mathrm{fc}}$ supervises foot-contact labels; and $\mathcal{L}_{\mathrm{skate}}$ penalizes foot sliding when the predicted contact label indicates ground contact but the joint velocity exceeds a threshold.

\paragraph{Global Stream Losses.}
The global stream is supervised by a similar set of physics-grounded losses:
\begin{equation}
\begin{split}
    \mathcal{L}_{\mathrm{traj}} = {} & \mathcal{L}_{\mathrm{FM}} + \lambda_{\mathrm{rec}} \mathcal{L}_{\mathrm{rec}} + \lambda_{\mathrm{rpos}} \mathcal{L}_{\mathrm{rpos}} \\
    & + \lambda_{\mathrm{rvel}} \mathcal{L}_{\mathrm{rvel}} + \lambda_{\mathrm{rsmooth}} \mathcal{L}_{\mathrm{rsmooth}}
\end{split}
\label{eq:traj_loss}
\end{equation}
The global stream is supervised similarly by $\mathcal{L}_{\mathrm{traj}} = \mathcal{L}_{\mathrm{FM}} + \lambda_{\mathrm{rec}} \mathcal{L}_{\mathrm{rec}} + \lambda_{\mathrm{rpos}} \mathcal{L}_{\mathrm{rpos}} + \lambda_{\mathrm{rvel}} \mathcal{L}_{\mathrm{rvel}} + \lambda_{\mathrm{rsmooth}} \mathcal{L}_{\mathrm{rsmooth}}$, where $\mathcal{L}_{\mathrm{rpos}} {=} \| \hat{\mathbf{J}}_0 - \mathbf{J}_0^{\mathrm{GT}} \|_2^2$, $\mathcal{L}_{\mathrm{rvel}} {=} \| \Delta_t \hat{\mathbf{J}}_0 - \Delta_t \mathbf{J}_0^{\mathrm{GT}} \|_2^2$, and $\mathcal{L}_{\mathrm{rsmooth}} {=} \| \Delta_t^2 \hat{\mathbf{J}}_0 \|_2^2$ are root position, velocity, and acceleration losses ($\mathbf{J}_0$: pelvis joint), and $\mathcal{L}_{\mathrm{rec}}$ is the trajectory representation MSE.

\section{Experiments}
\label{sec:experiments}

In this section, we validate three claims: (i)~OmniPoser performs on par with state-of-the-art methods across heterogeneous sparse-observation settings, (ii)~Flow Matching enables diversity in generation, and (iii)~cross-modal masking mechanism and Masked Conditioning Encoder is essential for generalization.

\subsection{Implementation Details}
\label{subsec:impl_details}

Networks are optimized independently with AdamW ($\beta_1{=}0.9$, $\beta_2{=}0.999$, wd${}=0.1$, lr${}=2{\times}10^{-4}$) using warmup-cosine scheduling (4K-step warmup) for 480K steps at batch size 256 with gradient clipping at norm 1.
Motion clips are sampled at 30\,fps ($T{=}33$ frames).
All training is conducted on a single NVIDIA A40 (48\,GB) and converges in ${\sim}3$~days.
At inference, we stick with 1 Euler ODE step, yielding $>$16$\times$ speedup over the 100+ diffusion steps of baselines (e.g., RoHM~\cite{Zhang2024RoHMRH}).

For more details, please refer to the supplementary material.

\subsection{Evaluation Protocols}
\label{subsec:datasets_metrics}

\subsubsection{Benchmarks and Baseline Methods}
To demonstrate the unified capability of OmniPoser as discussed in \cref{sec:intro}, we categorize different sensor configurations and baseline methods into four settings. 

\textbf{Setting 1 (Sparse Observation):} Only a sparse subset of joints is observed. We evaluate two schemes: (1) upper-body joint positions, evaluated on AMASS~\cite{Mahmood2019AMASSAO} and Nymeria~\cite{ma2024nymeria}; and (2) a random joint masking regime, evaluated on AMASS. The baseline methods are RoHM~\cite{Zhang2024RoHMRH} and HuMoR~\cite{Rempe2021HuMoR3H}.  AMASS evaluations use AMASS training, and Nymeria evaluation uses Nymeria training.

\textbf{Setting 2 (Egocentric Head 6D Pose):} Only head position and rotation are available from HMD signals. We compare against EgoEgo~\cite{Li2022EgoBodyPE} and EgoAllo~\cite{Yi2024EstimatingBA} across AMASS, RICH~\cite{huang2022rich}, and Nymeria.  AMASS and RICH evaluations use AMASS training, and Nymeria evaluation uses Nymeria training.

\textbf{Setting 3 (Mocap data from 6 IMUs with keypoint positions):} Six IMUs provide rotations at root and leaf joints, while upstream methods provide keypoint positions. We evaluate on AIST++~\cite{li2021aistplusplus}, TotalCapture~\cite{trumble2017totalcapture}, and Nymeria, comparing with Stereo-Inertial Poser (SIP)~\cite{tang2026stereoinertialposermetricaccurateshapeaware} and RobustCap~\cite{pan2023fusing}. AIST++ uses AMASS+AIST++ training, TotalCapture uses AMASS training, and Nymeria uses Nymeria training.

\textbf{Setting 4 (Real-world Evaluation):} We present qualitative visualization results on 3DPW~\cite{vonmarcard2018recovering} and Ego-Exo4D~\cite{Grauman2023EgoExo4DUS} to show performance in unconstrained outdoor and ego-centric scenes.  We use one cross-setting checkpoint to generate 3DPW and Ego-Exo4D examples. 

For dataset usage, we follow the standard splits: AMASS uses 15 sub-datasets for training and held-out sets for testing; Nymeria uses an 85:15 activity-stratified split; and RICH follows the EgoAllo split. 

For fair comparison, we train two versions of OmniPoser during evaluation: during \textbf{Ours-S} training, mask samples are drawn from the corresponding setting; during \textbf{Ours-U} training, mask samples are drawn from the masking configurations stated in \ref{subsec:masking}, i.e., predefined sensor layouts and real-world observability patterns. Since there are no unified baselines at present, we construct RoHM-U and RoHM++-U, which are based on current specialist SOTA, RoHM. RoHM-U retains the original RoHM architecture, but the mask samples drawn during training are identical to Ours-U. Meanwhile, we upgrade RoHM-U's architecture into \mbox{RoHM++-U}, stacking three setting-specific (namely, Setting 1,2,3) input encoders before the network. Baseline rows without \textbf{-U} indicate per-setting specialist models unless noted otherwise. Bold and underlined values indicate the best and second-best results, respectively.

For Nymeria, dashes denote unavailable metrics.
Nymeria dataset contains 260M frames with dense body MVNX annotations. Due to practical resource constraints throughout data preprocessing and large-scale training, we leave this for future work.

\subsubsection{Metrics}
We report MPJPE and PA-MPJPE (mm), Acceleration Error (m/s$^2$), Jerk (km/s$^3$), Per-vertex Error (mm), Contact Accuracy, Foot Skating (FS, mm), Ground Penetration (mm), Translation Error ($T_\text{head}$, mm), and Hips \& Shoulders Rotation Error (deg). In all tables in this section, $\downarrow$ means lower is better, while $\uparrow$ means higher is better.

\subsection{Quantitative Results}
\label{subsec:sota}

\subsubsection{Setting 1: Sparse Observation}
\label{subsubsec:setting1}

\paragraph{Setting 1: Sparse Observation (Upper-body joint positions).}
As introduced previously, this setting is intrinsically under-constrained because the entire lower body is unobserved.
Tab.~\ref{tab:setting1_upper_body} presents results for the upper-body setting. On AMASS, OmniPoser achieves 32.1\,mm MPJPE, outperforming both RoHM (34.8\,mm) and HuMoR (88.0\,mm).
HuMoR relies on expensive test-time optimization using a CVAE prior, struggling heavily with complete lower-body absence. RoHM (or RoHM++-U) addresses this using two cascaded conditional diffusion models explicitly trained for this masking pattern.
In contrast, Ours-S and Ours-U's performance are comparable to counterpart models, and surpasses at some of the metrics. We attribute this to the Masked Conditioning Encoder (MCE), which uses learnable embeddings to encode global structural priors for unobserved joints.
Furthermore, our model achieves the lowest skating ratio (0.031). This shows that our foot-skating loss (e.g., $\mathcal{L}_{\mathrm{skate}}$) generates valid foot-ground interactions even for invisible observations, at a fraction of the computational cost ($1$ ODE step vs.\ 100+ diffusion steps).
On Nymeria, our model reduces MPJPE by ${\sim}64\%$ over RoHM (21.1 vs.\ 59.3\,mm), alongside lower acceleration error, proving our learned priors transfer robustly in the wild.

\begin{table}[!htbp]
\centering
\caption{Quantitative results of Setting 1: Sparse Observation (Upper-body joint positions).}
\label{tab:setting1_upper_body}
\scriptsize
\setlength{\tabcolsep}{2pt}
\begin{tabular}{l ccccc cccc}
\toprule
& \multicolumn{5}{c}{AMASS} & \multicolumn{4}{c}{Nymeria (In-the-wild)} \\
\cmidrule(lr){2-6} \cmidrule(lr){7-10}
Method & JPE$\downarrow$ & Acc$\downarrow$ & Con$\uparrow$ & Ska$\downarrow$ & Pen$\downarrow$ & JPE$\downarrow$ & Acc$\downarrow$ & Con$\uparrow$ & Ska$\downarrow$ \\
\midrule
HuMoR & 88.0 & 3.3 & 0.68 & 0.230 & 2.59 & 88.5 & 8.26 & 0.78 & 0.685 \\
RoHM & \underline{34.8} & \textbf{2.3} & \textbf{0.95} & \underline{0.078} & \textbf{0.69} & \underline{59.3} & \underline{7.2} & \underline{0.91} & \underline{0.610} \\
\textbf{Ours-S} & \textbf{32.1} & \underline{2.7} & \underline{0.94} & \textbf{0.031} & \underline{1.84} & \textbf{21.1} & \textbf{1.39} & \textbf{0.95} & \textbf{0.033} \\
\midrule
RoHM-U & 158.7 & 2.9 & \underline{0.75} & 0.190 & \underline{1.48} & - & - & - & - \\
RoHM++-U & \underline{39.9} & \textbf{2.3} & \textbf{0.95} & \underline{0.081} & \textbf{0.56} & - & - & - & - \\
\textbf{Ours-U} & \textbf{36.7} & \underline{2.7} & \textbf{0.95} & \textbf{0.025} & 9.32 & - & - & - & - \\
\bottomrule
\end{tabular}
\end{table}

\paragraph{Setting 1: Sparse Observation (Random joint masking).}
Tab.~\ref{tab:setting1_random} evaluates a random masking regime where up to 15 out of 22 joints are randomly masked.
This setting explicitly tests the model's ability to handle the unstable signals typical of real-world sensors (as discussed in \cref{sec:intro}).
Ours-S achieves 35.7\,mm MPJPE. This advantage comes from our random masked training and the MCE's ability to preseve joint identity.  
Ours-U's slight worse metrics are acceptable, considering that our method does not introduce hand-crafted modules that differentiate across settings.

\begin{table}[!htbp]
\centering
\caption{Quantitative results of Setting 1: Sparse Observation (Random joint masking) on AMASS.}
\label{tab:setting1_random}
\scriptsize
\begin{tabular}{l cccc}
\toprule
Metric & HuMoR & RoHM & \textbf{Ours-S} & \textbf{Ours-U} \\
\midrule
JPE (mm) $\downarrow$               & 80.5  & \underline{38.2} & \textbf{35.7} & 49.3 \\
Accel. (m/s$^2$) $\downarrow$  & 3.3  & \textbf{1.9} & \underline{2.2} & 2.6   \\
Contact $\uparrow$              & 0.78  & \underline{0.95} & \textbf{0.97} & 0.93  \\
Skating $\downarrow$            & 0.184  & \underline{0.024} & \textbf{0.018} & 0.019 \\
\bottomrule
\end{tabular}
\end{table}

\subsubsection{Setting 2: Egocentric Head 6D Pose}
\label{subsubsec:setting2}
Only head position and rotation are available (HMD signal)---a longstanding ill-posed problem where the entire body below the neck must be generated from a single 6-DoF observation. We compare against EgoEgo~\cite{Li2022EgoBodyPE} and EgoAllo~\cite{Yi2024EstimatingBA}.

Tab.~\ref{tab:setting2_head_only} presents results across three benchmarks.
On AMASS, Ours-S achieves 83.2\,mm MPJPE, Ours-U achieve 86.2mm MPJPE, (${\sim}30\%$ improvement over EgoAllo's 119.7\,mm and $50\%$ over EgoEgo's 167.4\,mm).
EgoEgo couples prediction of global trajectory and local motion, which makes it perform not as well as our decoupled structure. EgoAllo uses special allocentric grounding and conditional diffusion designed purely for head-mounted setups, yielding accurate head alignment (6.2\,mm vs.\ our 17.1\,mm).
We consider our slightly higher head error a trade-off for OmniPoser's flexibility. Without hard-coding the head pose, our DiT's cross-attention propagates the 6-DoF HMD signal globally, allowing our method to outperform those baselines on full-body recovery.
On RICH, Ours-S achieves a $25\%$ improvement in MPJPE (131.6\,mm) compared to specialist methods. Ours-U exhibits a different set of trade-offs compared to RoHM++-U. 
The lower contact accuracy (0.71) is a shared degradation across all methods due to inherent floor-height ambiguities in RICH unseen during AMASS training.
On Nymeria, OmniPoser reduces MPJPE by ${\sim}60\%$ over EgoAllo (40.7 vs.\ 103.0\,mm). Its superior contact accuracy (0.94 vs.\ EgoAllo's 0.63) highlights that our physics-grounded regularization term ensures stability even for unconstrained ego-centric capture.

\begin{table}[t]
\centering
\caption{Quantitative results of Setting 2: Egocentric Head 6D Pose.}
\label{tab:setting2_head_only}
\scriptsize
\setlength{\tabcolsep}{1.8pt}
\begin{tabular}{l cccc cccc ccc}
\toprule
& \multicolumn{4}{c}{AMASS} & \multicolumn{4}{c}{RICH (Cross-domain)} & \multicolumn{3}{c}{Nymeria (Wild)} \\
\cmidrule(lr){2-5} \cmidrule(lr){6-9} \cmidrule(lr){10-12}
Method & JPE$\downarrow$ & PA$\downarrow$ & Con$\uparrow$ & $T_h\downarrow$ & JPE$\downarrow$ & PA$\downarrow$ & Con$\uparrow$ & $T_h\downarrow$ & JPE$\downarrow$ & PA$\downarrow$ & Con$\uparrow$ \\
\midrule
EgoEgo & 167.4 & 145.8 & \underline{0.92} & 54.9 & 207.8 & 187.8 & \underline{0.88} & 66.5 & 135.8 & 117.2 & 0.57 \\
EgoAllo & \underline{119.7} & \underline{101.1} & \textbf{1.00} & \textbf{6.2} & \underline{176.2} & \underline{160.1} & \textbf{0.96} & \textbf{8.9} & \underline{103.0} & \underline{89.3} & \underline{0.63} \\
\textbf{Ours-S} & \textbf{83.2} & \textbf{73.5} & \underline{0.92} & \underline{17.1} & \textbf{131.6} & \textbf{111.5} & 0.71 & \underline{25.7} & \textbf{40.7} & \textbf{36.2} & \textbf{0.94} \\
\midrule
RoHM-U & 124.3 & 101.5 & \underline{0.94} & 36.7 & 195.0 & 115.4 & 0.62 & 56.8 & - & - & - \\
RoHM++-U & \underline{94.5} & \underline{92.5} & \textbf{0.96} & \textbf{19.5} & \textbf{126.3} & \underline{112.3} & \textbf{0.72} & \textbf{44.8} & - & - & - \\
\textbf{Ours-U} & \textbf{86.2} & \textbf{69.8} & 0.92 & \underline{33.2} & \underline{134.8}& \textbf{106.4} & \underline{0.65} & \underline{50.7} & - & - & - \\
\bottomrule
\end{tabular}
\end{table}

\subsubsection{Setting 3: Mocap Data from 6 IMUs with Keypoint Positions}
\label{subsubsec:setting3}
Six IMUs at root and leaf joints (pelvis, head, wrists, ankles) provide rotations, while upstream methods (i.e. pose estimation results from RGB images, odometry sensor readouts) provide keypoint positions. We compare with Stereo-Inertial Poser (SIP)~\cite{tang2026stereoinertialposermetricaccurateshapeaware} and RobustCap~\cite{pan2023fusing}. We note that the original RobustCap leverages 33 MediaPipe 2D keypoints from an external camera (as camera-normalized marching rays), while OmniPoser uses IMU orientations together with 3D joint positions in the inertial frame. To ensure a fair comparison, we enhance and retrain RobustCap by replacing its 2D keypoint inputs with 3D keypoints matching OmniPoser's input representation, denoted \textbf{RobustCap3D}; this upgrade should improve its global translation estimation while providing a controlled ablation of input modality.

\paragraph{AIST++.}
On the highly dynamic AIST++ dataset, Ours-S / Ours-U achieves the best PA-MPJPE (19.4\,mm / 21.4\,mm) and MPJPE (26.0\,mm / 29.3\,mm).
Crucially, the Ours-S's jerk metric (0.29\,km/s$^3$) is $5{\times}$ lower than SIP and $8{\times}$ lower than RobustCap. Unlike RobustCap, which directly regresses visual and inertial cues leading to high-frequency jitter, OmniPoser leverages a decoupled generative architecture. The temporal convolutional UNet smoothly filters the low-frequency global trajectory, which then conditions the DiT to suppress local kinematic jitter, assisted strongly by our velocity and smoothness losses.

\paragraph{TotalCapture.}
On the TotalCapture benchmark, Ours-S outperforms RobustCap3D in MPJPE (36.2 vs.\ 42.3\,mm), and Ours-U stays competitive with RoHM++-U across metrics.
Logically, SIP achieves superior global translation (5.42 vs.\ 15.9\,mm) and hip/shoulder rotation ($10.6^{\circ}$ vs.\ $13.3^{\circ}$) due to the fact that it extracts absolute metric depth from stereo cameras and directly substitutes physical pelvis IMU readings for root orientation.
In contrast, OmniPoser strictly \emph{generates} global trajectory and local rotations from masked observations using general flow matching. The slight performance gap in rigid root translation is a conscious trade-off to preserve a purely modality-agnostic formulation.

\paragraph{Nymeria.}
On Nymeria, OmniPoser's lower MPJPE (16.6 mm) and jerk (0.048 m/s$^3$) reinforce that our cross-modal masking conditioning effectively learns unified kinematic priors, avoiding the need to tune multi-modal sensor fusion heuristics for real-world deployment.

\begin{table*}[t]
\centering
\caption{Quantitative results of Setting 3: Mocap Data from 6 IMUs with Keypoint Positions.}
\label{tab:setting3_6imu_with_kpt_pos}
\resizebox{\textwidth}{!}{
\setlength{\tabcolsep}{1.1pt}
\begin{tabular}{l ccccccc ccccccc ccccccc}
\toprule
& \multicolumn{7}{c}{AIST++} & \multicolumn{7}{c}{TotalCapture} & \multicolumn{7}{c}{Nymeria (In-the-wild)} \\
\cmidrule(lr){2-8} \cmidrule(lr){9-15} \cmidrule(lr){16-22}
Method & PA$\downarrow$ & JPE$\downarrow$ & PVE$\downarrow$ & TE$\downarrow$ & HS$\downarrow$ & Jrk$\downarrow$ & FS$\downarrow$ & PA$\downarrow$ & JPE$\downarrow$ & PVE$\downarrow$ & TE$\downarrow$ & HS$\downarrow$ & Jrk$\downarrow$ & FS$\downarrow$ & PA$\downarrow$ & JPE$\downarrow$ & PVE$\downarrow$ & TE$\downarrow$ & HS$\downarrow$ & Jrk$\downarrow$ & FS$\downarrow$ \\
\midrule
SIP & 23.1 & 29.8 & 39.0 & \textbf{11.31} & \underline{8.91} & \underline{1.50} & \underline{0.57} & 28.6 & 46.5 & 54.3 & \textbf{5.42} & \textbf{10.6} & \underline{0.94} & \textbf{0.69} & 15.2 & \underline{17.0} & \underline{16.8} & \textbf{4.13} & \textbf{5.22} & \underline{0.059} & \textbf{0.25} \\
RobCap & 24.0 & 33.1 & 43.2 & 19.92 & 9.34 & 2.49 & 0.61 & 33.5 & 48.7 & 63.4 & 23.5 & 13.4 & 2.09 & 1.40 & 16.5 & 19.5 & 19.4 & 4.52 & 7.51 & 1.56 & 0.54 \\
RobCap3D & \underline{21.8} & \underline{28.5} & \underline{37.6} & \underline{15.82} & 8.95 & 2.31 & 0.59 & \underline{27.9} & \underline{42.3} & \underline{52.8} & \underline{9.74} & \underline{12.8} & 1.87 & 1.22 & \underline{15.0} & 17.8 & 17.5 & \underline{4.25} & \underline{7.18} & 1.41 & 0.48 \\
\textbf{Ours-S} & \textbf{19.4} & \textbf{26.0} & \textbf{32.9} & 16.54 & \textbf{7.40} & \textbf{0.29} & \textbf{0.22} & \textbf{26.5} & \textbf{36.2} & \textbf{51.2} & 15.9 & 13.3 & \textbf{0.16} & \underline{0.81} & \textbf{14.5} & \textbf{16.6} & \textbf{16.6} & 6.83 & 7.50 & \textbf{0.048} & \underline{0.39} \\
\midrule
RoHM-U & 123.5 & 157.5 & 185.9 & 76.3 & 38.1 & \textbf{0.21} & \underline{9.49} & 78.6 & 95.3 & 100.4 & 22.8 & 19.5 & 0.19 & \underline{0.95} & \multicolumn{7}{c}{-} \\
RoHM++-U & \underline{89.5} & \underline{136.5} & \underline{128.5} & \underline{56.1} & \underline{32.1} & \textbf{0.21} & 9.51 & \textbf{18.7} & \textbf{29.0} & \textbf{26.3} & \underline{17.5} & \textbf{8.3} & \textbf{0.07} & 2.06 & \multicolumn{7}{c}{-} \\
\textbf{Ours-U} & \textbf{21.4} & \textbf{29.3} & \textbf{40.4} & \textbf{17.5} & \textbf{7.9} & \underline{1.73} & \textbf{1.05} & \underline{23.4} & \underline{35.5} & \underline{47.2} & \textbf{15.7} & \underline{11.1} & \underline{0.15} & \textbf{0.86} & \multicolumn{7}{c}{-} \\
\bottomrule
\end{tabular}
}
\end{table*}

\subsection{Qualitative Results}
\label{subsec:qualitative}
\begin{figure*}[!t]
\centering
\includegraphics[width=\linewidth]{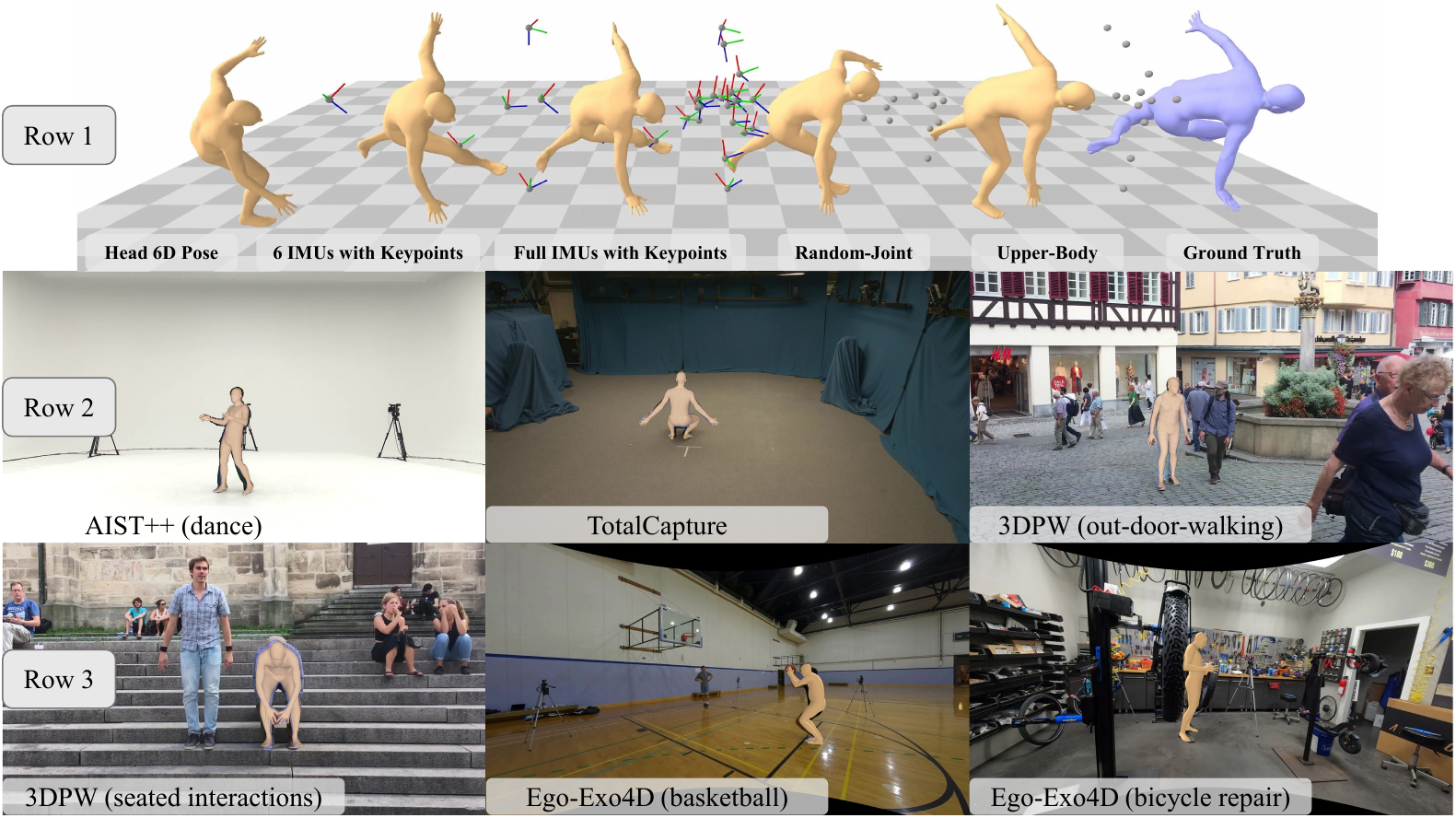}
\caption{%
\textbf{Qualitative results.}
\emph{Row 1:} Five masking configurations applied to the same speed-vault sequence from AMASS dataset. Meshes in gold are the predictions. The mesh on the rightmost is the ground truth.
\emph{Row 2--3:} Visualizations on different datasets.
For more qualitative results, please refer to the supplementary material.
}
\label{fig:qualitative_grid}
\end{figure*}

Importantly, all qualitative results in this section are produced by Ours-U model~\ref{subsec:datasets_metrics}.
In the first row in Fig.~\ref{fig:qualitative_grid}, we demonstrate this capability: given a speed-vault clip, one model produces plausible reconstructions under five distinct masking configurations at inference time. Configurations with spatially distributed observations (6 IMUs with keypoints, full IMUs with keypoints) most faithfully track the ground truth, while the head 6d pose setting anchors the head 6-DoF pose and recovers a plausible full-body pose. The upper-body setting, lacking any lower-body supervision, produces a less dynamic but still coherent trajectory.
Fig.~\ref{fig:qualitative_grid} (Row 2--3) shows faithful recovery on dynamic AIST++ dances and tight alignment on TotalCapture.
On 3DPW, OmniPoser produces coherent walking gaits and seated interactions in outdoor scenes, validating that kinematic priors learned from controlled mocap dataset generalize to in-the-wild domains.
On Ego-Exo4D, OmniPoser recovers whole-body motion from ego-centric basketball and bicycle-repair activities involving complex body-object interactions, showing our method's applicability to AR/VR capture scene.

\subsection{Ablation Studies}
\label{subsec:ablation}

We ablate OmniPoser from two aspects: the effectiveness of MCE module, and the sampling diversity from our generative conditional network.

\begin{table}[t]
\centering
\caption{\textbf{Ablation studies} (AMASS, upper-body).}
\label{tab:ablation}
\scriptsize
\setlength{\tabcolsep}{4pt}
\begin{tabular}{l ccc}
\toprule
Variant & JPE$\downarrow$ & Acc$\downarrow$ & Ska$\downarrow$ \\
\midrule
\textbf{OmniPoser (full)} & \textbf{32.1} & \textbf{2.7} & \textbf{0.031} \\
\midrule
MCE $\to$ zero-padding & 88.1 & 4.0 & 0.053 \\
\bottomrule
\end{tabular}
\end{table}

\begin{figure}[!htbp]
\centering
\begin{minipage}[t]{1\linewidth}
\centering
\includegraphics[width=\linewidth]{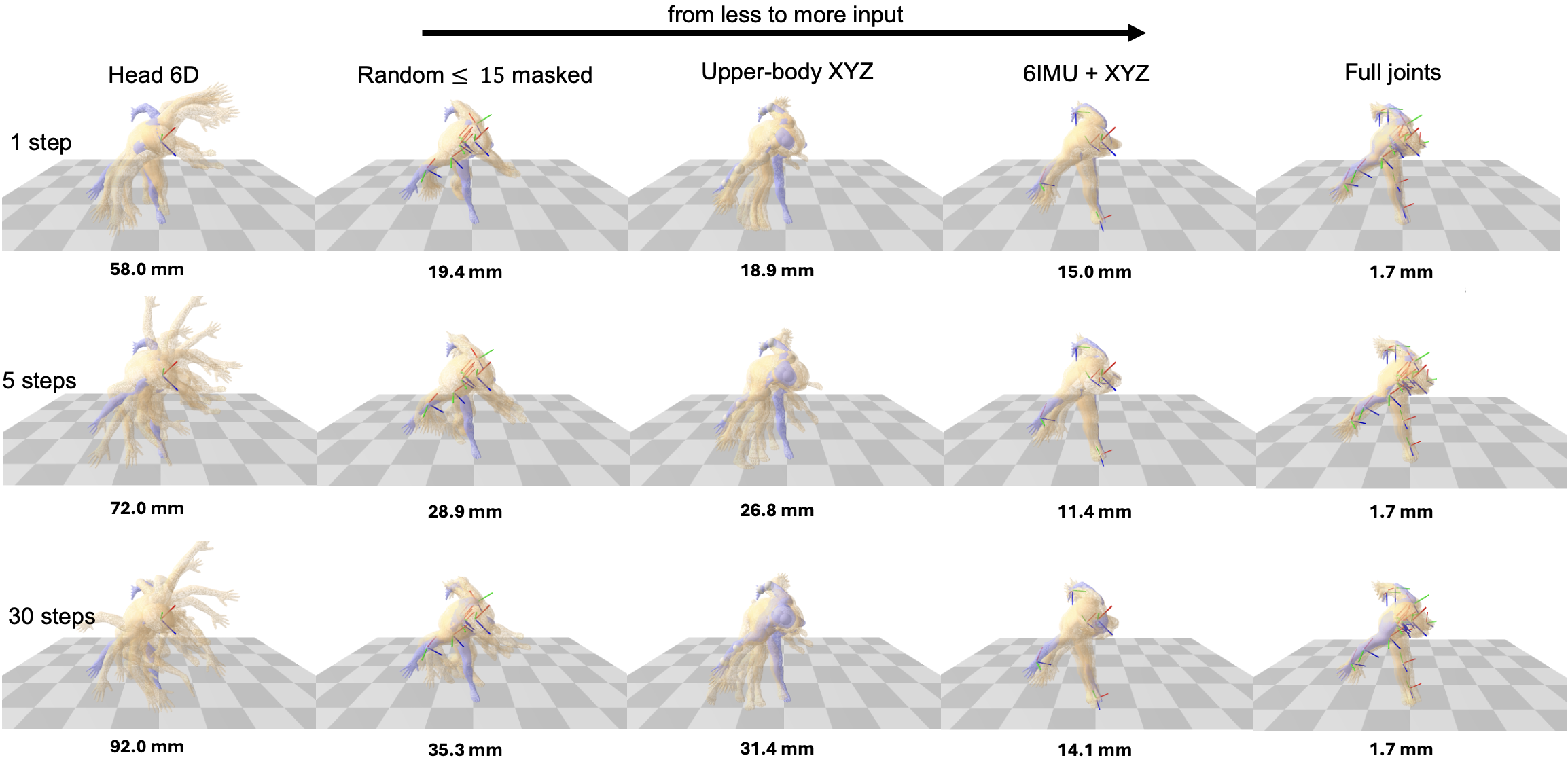}
\caption{Diversity under input constraints and ODE steps.}
\label{fig:diversity_input_constraints}
\end{minipage}
\hfill
\begin{minipage}[t]{1\linewidth}
\centering
\captionof{table}{Sampling diversity, measured as joint-position standard deviation in mm over 100 samples.}
\label{tab:diversity_steps}
\scriptsize
\setlength{\tabcolsep}{3.5pt}
\resizebox{\linewidth}{!}{
\begin{tabular}{l ccccc}
\toprule
ODE steps & Head 6D & Random $\leq$15 masked & Upper-body XYZ & 6IMU rot.+XYZ & Full joints \\
\midrule
1  & 58.0 & 19.4 & 18.9 & 15.0 & 1.7 \\
5  & 72.0 & 28.9 & 26.8 & 11.4 & 1.7 \\
30 & 92.0 & 35.3 & 31.4 & 14.1 & 1.7 \\
\bottomrule
\end{tabular}
}
\end{minipage}
\end{figure}

\subsubsection{MCE learnable embeddings.}
Replacing per-joint learnable embeddings with zero-padding weakens the model's ability to distinguish \emph{which} joints are missing, worsening MPJPE by $+$56.0\,mm.

\subsubsection{Diversity, input constraints, and inference steps.}
We sample 100 predictions for the same specific motion while varying only the observed mask and Euler ODE steps. 
The std of joint positions and 10 overlaid samples are shown in the Fig.~\ref{fig:diversity_input_constraints}. Tab.~\ref{tab:diversity_steps} show how sampling diversity varies according to input sensor configurations and flow matching sampling steps.
Diversity grows as input signals decrease (rows) and as steps increase (columns), demonstrating that our model can output human motion generatively while conforming to input constraints. Quantitatively, however, we hardly observe any performance increase with more steps, so we use single-step for benchmarking. In real-world applications, users can trade speed for diversity (5 steps @ 250 fps, 30 steps @ 80 fps).

\section{Conclusion}

OmniPoser treats various flexible inputs (e.g. partial keypoints, head-mounted 6-DoF poses, sparse IMUs) as partial views of one shared body model, making it unified for different settings.

We achieve this through three components: modular masking with a Masked Conditioning Encoder, a decoupled dual-stream architecture, and Conditional Flow Matching, which altogether allow one model to cover diverse sensor setups while enabling real-time, single-step ODE inference.
Across three settings and seven benchmarks, OmniPoser consistently performs on par with state-of-the-art methods while running at over 700\,FPS.
However, several limitations remain.
First, OmniPoser models a single person and does not account for body-object or body-body contact forces; extending the masking abstraction to multi-person interaction or contact-rich object manipulation is an active direction for future work.
Second, the cascaded inference pipeline propagates trajectory errors into pose recovery with no built-in correction mechanism; incorporating iterative refinement or feedback between the two streams could further improve robustness, though in practice the single-pass cascade already produces temporally smooth and drift-free trajectories across all evaluated benchmarks.
Third, OmniPoser currently operates on the SMPL 22-joint skeleton and does not support hand or face articulation (SMPL-X); the modular feature-group design, however, is readily extensible to finer-grained body models as the corresponding training data becomes available.
Extending the masking abstraction to multi-person interaction, contact-rich object manipulation, and articulated hand/face models would adapt our framework to a broader set of embodied tasks.

\section*{Acknowledgements}
This work was supported by Fundamental and Interdisciplinary Disciplines Breakthrough Plan of the Ministry of Education of China, the Shanghai Commitee of Science and Technology, China (Grant No.24511103200), Shanghai Artificial Intelligence Laboratory, and XPLORER PRIZE grants.

\bibliographystyle{splncs04}
\bibliography{main}

\end{document}